\documentclass[
]{ceurart}

\usepackage{listings}
\usepackage{svg} 
\begin{document}

\copyrightyear{2023}
\copyrightclause{Creative Commons License Attribution 4.0
  International (CC BY 4.0).}

\conference{TOTh 2023: Terminology \& Ontology: Theories and applications June 1--2, 2023, Chambery, France}

\title{QICHWABASE: A Quechua Language and Knowledge Base for Quechua Communities}

\author[1]{Elwin Huaman}[%
orcid=0000-0002-2410-4977,
email=Elwin.Huaman@uibk.ac.at,
]
\cormark[1]
\address[1]{University of Innsbruck, Austria}

\author[2]{David Lindemann}[%
orcid=0000-0002-8261-6882,
email= david.lindemann@ehu.eus,
]
\address[2]{University of the Basque Country, Spain}

\author[3]{Valeria Caruso}[%
email=vcaruso@unior.it,
]
\address[3]{University of Naples L'Orientale, Italy}

\author[4]{Jorge Luis Huaman}[%
email=jorgellhq@gmail.com,
]
\address[4]{National University of the Altiplano of Puno, Peru}

\begin{abstract}
Over the last decade, the Web has increasingly become a space of language and knowledge representation. However, it is only true for well-spread languages and well-established communities, while minority communities and their resources received less attention.
In this paper, we propose QICHWABASE to support the harmonization process of the Quechua language and knowledge, and its community. For doing it, we adopt methods and tools that could become a game changer in favour of Quechua communities around the world. 
We conclude that the methodology and tools adopted on building QICHWABASE, which is a Wikibase instance, could enhance the presence of minorities on the Web.
\end{abstract}


\maketitle

\section{Introduction}
The availability of interoperable linguistics resources is nowadays more urgent in order to save and help under-resourced languages, and their communities. Despite the efforts, not all languages are represented on the Web, nor made accessible in a structured format (e.g. Knowledge Graph).
Well-spread languages, like English or Spanish, took up an overwhelming majority on the Web~\cite{AghaeiAHBSF22}, while indigenous communities and their resources received less attention, e.g., there is no structured knowledge base dedicated to the Quechua community.

\begin{figure}
  \centering
  \includegraphics[width=\linewidth]{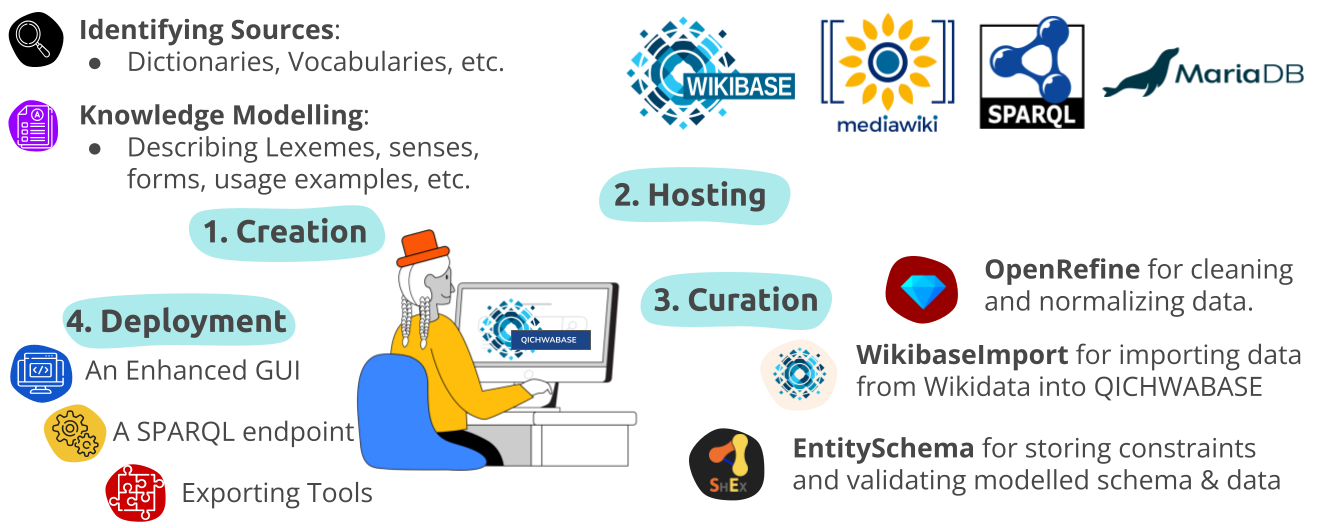}
  \caption{QICHWABASE generation workflow: creation, hosting, curation, and deployment.}
    \label{fig:qichwabase}
\end{figure}

This paper presents our ongoing work on the QICHWABASE\footnote{\url{https://qichwa.wikibase.cloud/}}~\cite{HuamanHH2022}, which aims to support a harmonization process of the Quechua language and knowledge, and the Quechua community.
The main contribution of this paper is the creation of a dedicated knowledge base for the Quechua community. To do it, we followed a process model for knowledge graph generation~\cite{FenselSAHKPTUW20}, which involves: knowledge creation, knowledge hosting, knowledge curation~\cite{HuamanF2022}, and knowledge deployment (see Figure~\ref{fig:qichwabase}).

To date, QICHWABASE (a Wikibase\footnote{\url{https://wikiba.se/}} instance) is accessible on the Web, retrievable by machines, and curated by the Quechua community, and it contains 1 030 825 triples (or statements), lexemes in various Quechua dialects, and sense descriptions in English, German, and Spanish.

\section{Implementation and Use Cases}

We build the QICHWABASE based on the phases described in Figure \ref{fig:qichwabase}. Then, we hosted it on Wikibase Cloud\footnote{\url{https://wikibase.cloud/}}, which provides a full-fledged and fully managed Wikibase infrastructure. Afterwards, we analyse the factors that may compromise the success of QICHWABASE: i) No dedicated tools to support tasks in the overall QICHWABASE workflow, and ii) Lack of an active Quechua community on the Web. For the first challenge, Wikibase offers various services and integration with external tools that are continuously maintained by the community. For the second challenge, we are leading various talks, tutorials, and dissemination in order to gather the Quechua community to be part of QICHWABASE. Finally, some use case scenarios can be mentioned as follows: i) Answering (subject, predicate, object) questions, ii) improving and supporting entity linking and knowledge validation~\cite{HuamanKF2020} tasks, and iii) supporting the development of applications, such as dialogue systems, chatbots, etc.

\section{Conclusion and Future Work}

In this paper, we have described our effort for building QICHWABASE from scratch. We aim to achieve a collaborative mentality and participation throughout the quechua community, so a high-quality knowledge base can be built, maintained, used, and shared for improving several use case scenarios.
In the short term, we will continue ingesting new sources into QICHWABASE and run workshops and tutorials in order to integrate the participation of Quechua communities around the world in the QICHWABASE life cycle.
\begin{acknowledgments}
  We would like to thank the NexusLinguarum COST Action CA18209.  
\end{acknowledgments}

\bibliography{bib}

\appendix

\section{Poster}

\begin{figure}[h]
  \centering
  \includegraphics[width=0.80\textwidth]{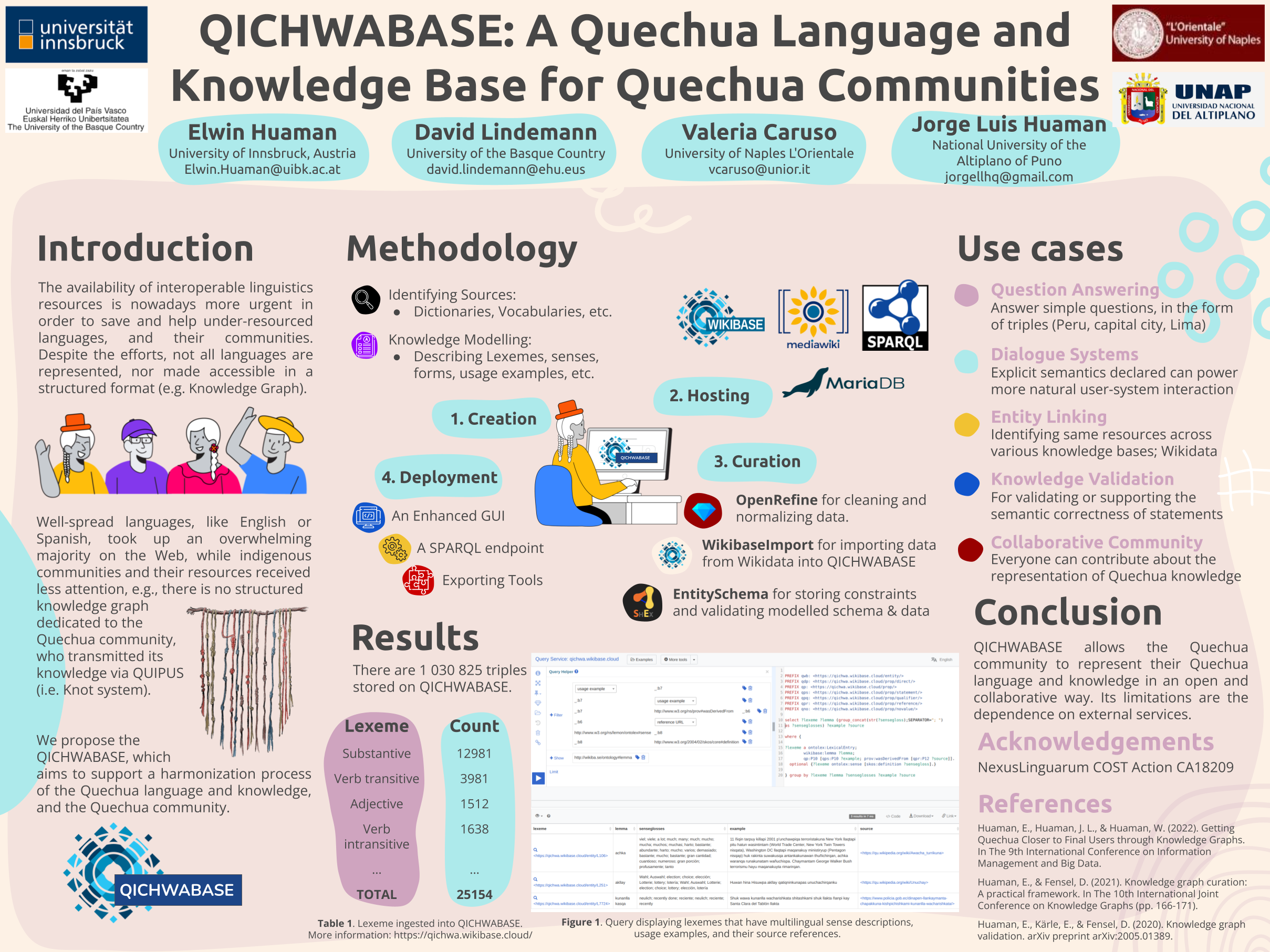}
  \caption{Poster | QICHWABASE: A Quechua Language and Knowledge Base for Quechua Communities.}
    \label{fig:poster}
\end{figure}

\end{document}